\documentclass[lettersize,journal]{IEEEtran}
\usepackage{amsmath,amsfonts}
\usepackage{algorithmic}
\usepackage{algorithm}
\usepackage{array}
\usepackage[caption=false,font=normalsize,labelfont=sf,textfont=sf]{subfig}
\usepackage{textcomp}
\usepackage{stfloats}
\usepackage{url}
\usepackage{verbatim}
\usepackage{graphicx}
\usepackage{cite}
\usepackage[colorlinks=true, linkcolor=blue, citecolor=blue, urlcolor=blue]{hyperref}

\usepackage{enumitem}
\usepackage{graphicx}
\usepackage{epstopdf}
\usepackage{multirow}
\usepackage{bbding}
\usepackage{pifont}
\usepackage{wasysym}
\usepackage{amssymb}
\usepackage{float}
\usepackage[table]{xcolor}
\usepackage{colortbl}

\begin{document}

\title{3UR-LLM: An End-to-End Multimodal Large Language Model for 3D Scene Understanding}

\author{Haomiao Xiong,  Yunzhi Zhuge, Jiawen Zhu, Lu Zhang, Huchuan Lu,~\IEEEmembership{Fellow~IEEE}
\thanks{Manuscript received 11 October 2024; revised 4 December 2024; accepted 4 December 2024. (Corresponding author: Yunzhi Zhuge.)}
\thanks{Haomiao Xiong,  Yunzhi Zhuge, Jiawen Zhu, Lu Zhang are with the School of Information and Communication Engineering, Dalian University of Technology, Dalian 116081, China. (e-mail: 22309071@mail.dlut.edu.cn; zgyz@dlut.edu.cn; jiawen@mail.dlut.edu.cn;  zhangluu@dlut.edu.cn)}
\thanks{Huchuan Lu is with the School of Future Technology and the School of Artificial Intelligence, Dalian University of Technology, Dalian 116081, China. (email: lhchuan@dlut.edu.cn)}}

\markboth{IEEE Transactions on Multimedia}%
{Shell \MakeLowercase{\textit{et al.}}: A Sample Article Using IEEEtran.cls for IEEE Journals}


\maketitle

\begin{figure*}
  \centering
  \setlength{\abovecaptionskip}{0cm}
  \includegraphics[width=0.975\textwidth]{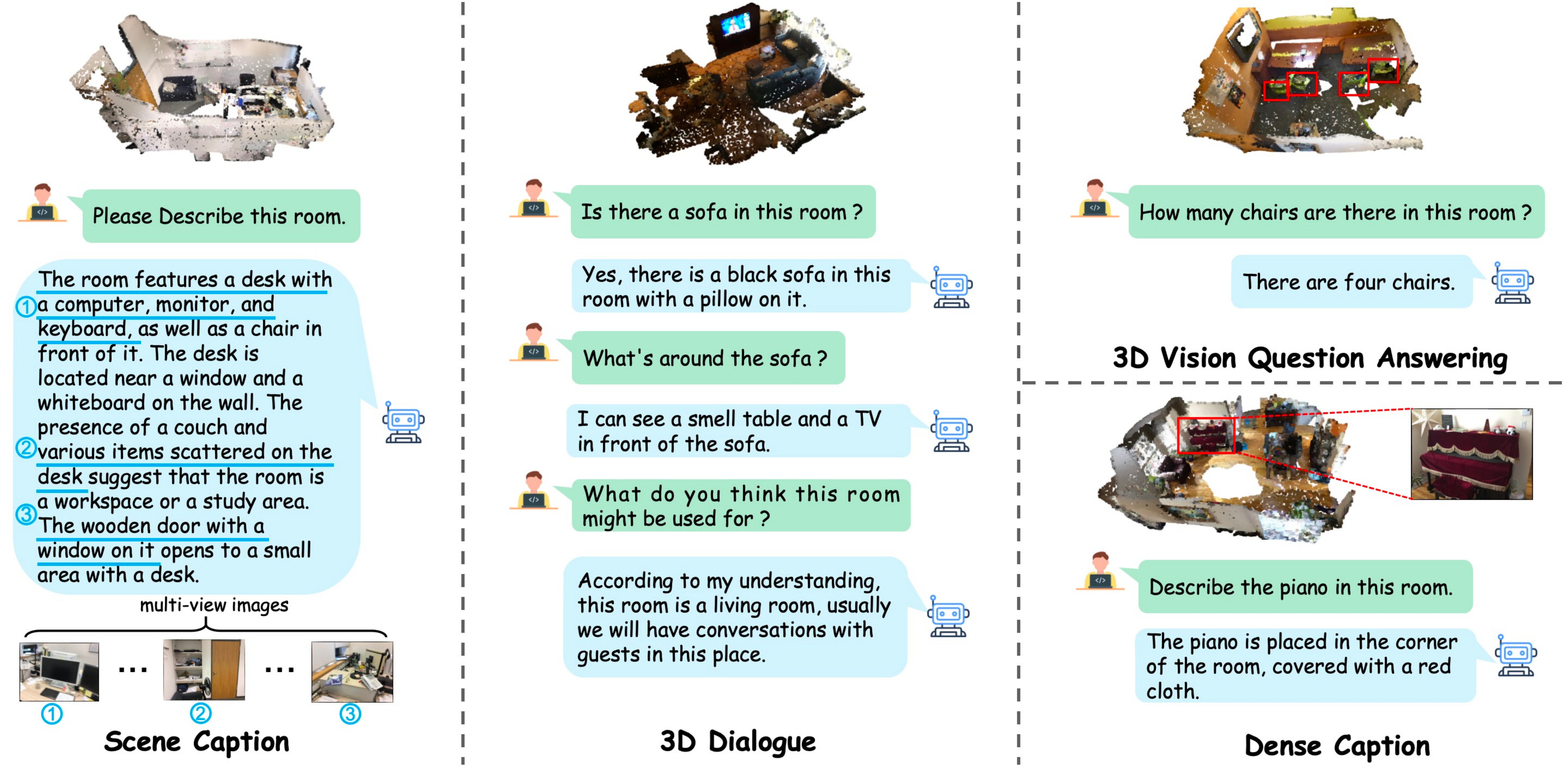}
  \vspace{-2mm}
  \caption{Demonstration of 3UR-LLM, a multi-modal large language model tailored for comprehending 3D scenes, which enables flexible and precise responses to given instructions, showing robust reasoning capabilities in real-world environments. 
  }
  \vspace{-2mm}
  \label{figure:3UR-LLM}
\end{figure*}

\begin{abstract}
Multi-modal Large Language Models (MLLMs) exhibit impressive capabilities in 2D tasks, yet encounter challenges in discerning the spatial positions, interrelations, and causal logic in scenes when transitioning from 2D to 3D representations.
We find that the limitations mainly lie in: 
\romannumeral1) the high annotation cost restricting the scale-up of volumes of 3D scene data, and 
\romannumeral2) the lack of a straightforward and effective way to perceive 3D information which results in prolonged training durations and complicates the streamlined framework. 
To this end, we develop a pipeline based on open-source 2D MLLMs and LLMs to generate 
high-quality 3D-text pairs and construct 3DS-160K 
, to enhance the pre-training process. 
Leveraging this high-quality pre-training data, we introduce the 3UR-LLM model, an end-to-end 3D MLLM designed for precise interpretation of 3D scenes, showcasing exceptional capability in navigating the complexities of the physical world. 
3UR-LLM directly receives 3D point cloud as input and project 3D features fused with text instructions into a manageable set of 
tokens.
Considering the computation burden derived from these hybrid tokens, 
we design a 3D compressor module
to cohesively
compress the 3D spatial cues and textual narrative.
3UR-LLM achieves promising performance with respect to the previous SOTAs, 
for instance, 3UR-LLM exceeds its counterparts by 7.1\% CIDEr on ScanQA, while utilizing fewer training resources.
The code and model weights for 3UR-LLM and the 3DS-160K benchmark are available at  \href{https://github.com/hmxiong/3UR-LLM}{3UR-LLM}.

\end{abstract}


\section{Introduction}

Recent advances in Large Language Models (LLMs), such as ChatGPT~\cite{openai2022chatgpt}, LLaMA~\cite{touvron2023llama,dubey2024llama}, and Gemini~\cite{team2023gemini}, have shown impressive capabilities in tasks, $e.g.$ language understanding and generation. 
These advancements have facilitated the utilization of LLMs' reasoning abilities for visual tasks, leading to the rise of Multi-modal Large Language Models (MLLMs)~\cite{li2023blip-2, awadalla2023openflamingo, liu2023improved, panagopoulou2023x}. 
These models have significantly promoted downstream tasks such as scene captioning, visual question answering, and dense perception tasks. 
However, while 2D visual tasks have greatly benefited from extensive research, 3D scene understanding presents more complex challenges such as spatial perception, geometric parsing, and the comprehension of 
inter-object relationships and interactions within the 3D physical environments.
Consequently, integrating the world knowledge captured by MLLMs into 3D scene understanding presents a promising yet formidable frontier that demands further exploration.


Recent progress in 3D scene understanding paralleled the complexities of interpreting the physical world, from 3D spatial comprehension as detailed in ScanQA~\cite{azuma2022scanqa} to situated question answering as shown in SQA3D~\cite{ma2022sqa3d}.  They employed specialized text encoders, $e.g.$ BERT~\cite{devlin2018bert}, and 3D visual encoders, $e.g.$ VoteNet~\cite{qi2019deep}, to parse multi-modal inputs and explore feature fusion strategies.
Building on this,  3D-LLM~\cite{hong20233d} illustrating the potential of MLLMs in 3D tasks by leveraging the advanced reasoning of LLMs. 


Despite notable advancements, current 3D scene understanding methods often face challenges with adaptability in query response, as AI agents in the 3D world are typically bound by task-specific designs.  This limitation restricts their ability to flexibly interpret human questions and provide detailed responses, relying instead on selecting predefined answers through classification mechanisms~\cite{azuma2022scanqa,ma2022sqa3d}.
Furthermore, integrating 3D data into these models introduces additional challenges. For instance, the 3D-LLM framework~\cite{hong20233d} does not process 3D point clouds directly but depends on separate algorithms~\cite{kirillov2023segment, cheng2022masked} to convert multi-view images into 3D embeddings, which increases complexity and response time.
There is also a critical lack of high-quality, instruction-based 3D-text data for training these models. Unlike the easily scalable 2D multi-modal data available online, the rarity of similar 3D datasets hampers advancement.  Attempts to create 3D-text pairs with advanced LLMs like ChatGPT~\cite{ouyang2022training} face substantial obstacles due to the high costs and limitations associated with closed-source models, further restricting dataset expansion and innovation in 3D scene understanding.

To address the aforementioned challenges, in this work, we introduce a novel end-to-end architecture, termed 3UR-LLM, that formulates the problem of 3D scene understanding by conceptualizing it as the interpretation of multi-modal environments and language generation of response to human instructions.

3UR-LLM is designed to facilitate reasoning and sustain continuous dialogues within 3D environments,  mirroring 
intelligent human-agent interactions and enabling agents to engage in further planning with a detailed perceptual understanding of the environment. Demonstrated in Fig.~\ref{figure:3UR-LLM}, 3UR-LLM excels in generating relevant and detailed responses to diverse queries about the environment, facilitating comprehensive dialogue and planning capabilities.



Specifically, establishing a comprehensive perception within the 3D environment 
serves as an essential foundation for LLMs to produce relevant results. 
Different from 3D-LLM~\cite{hong20233d}  which resorts 
to 
reconstruction algorithms to derive 3D representations from 2D multi-view
images, 
we directly integrate 3DETR~\cite{misra2021end} as our perception component. This strategy not only
simplifies our framework but also leverages 3DETR's object-level 3D structural insights to better grasp the intricate relationships.
Furthermore, we introduce a 3D compression module to streamline and align 3D features. 
Inspired by the multi-modal transformer in BLIP~\cite{li2022blip}, our 3D compressor projects extensive 3D features into a controllable set of vision tokens, optimizing computation efficiency during training and inference. 
Additionally, to enhance the spatial understanding of the compressor, we design a 3D query fusion mechanism to select high-confidence queries 
from perception component 
for joint training with the original queries from the compressor. 
These innovations endow 3UR-LLM to robustly perceive 3D environments and produce precise predictions.


To tackle the challenges 
concerning
3D-text modality alignment
training data, we leverage open-source LLaVA~\cite{liu2023improved} and Vicuna~\cite{chiang2023vicuna} as agents to establish a data engine 
that generates
the desired 3D-text pairs. 
Through meticulously designed prompts and a systematic process of generation and refinement, we obtain 160K high-quality 3D-text samples based on multi-view images from 
\cite{dai2017scannet, wald2019rio}, which we term as 3DS-160K. 
It covers a wide range of 3D physical scenes and is conducive to various tasks such as 3D scene descriptions, dense captions, and question answering, thus significantly enhancing the data available for 3D scene understanding research.

In summary, our main contributions are as follows:
\begin{itemize} [leftmargin=0.1cm, itemindent=0.5cm]
    \item{
    We launch 3UR-LLM, a novel multimodal large language model tailored for end-to-end 3D scene understanding. This model integrates 3D point clouds with linguistic features, processed through a multimodal transformer to model and understand the 3D world effectively.}
    \item{
    We develop a 3D compression module that efficiently projects 3D features into a condensed set of vision tokens. This module works in tandem with a 3D query fusion mechanism that integrates high-confidence queries to enhance spatial understanding, significantly improving computational efficiency and robustness in 3D environment perception.
    }
    \item{
    A data engine leveraging open-source MLLM and LLM technologies has been constructed to annotate datasets. Additionally, we create the 3DS-160K dataset, a rich collection of 160,000 high-quality 3D-text pairs, designed to support a variety of tasks including 3D scene description, dense captioning, and question answering.
    }
    \item{    
    Extensive experiments demonstrate the effectiveness of our model, which not only surpasses existing approaches but also provides a more efficient pipeline, reducing the resources required for training.
    }
\end{itemize}

\section{Related Work}

\begin{figure*}[htbp]
\vspace{-1mm}
\centering
\includegraphics[scale=0.215]{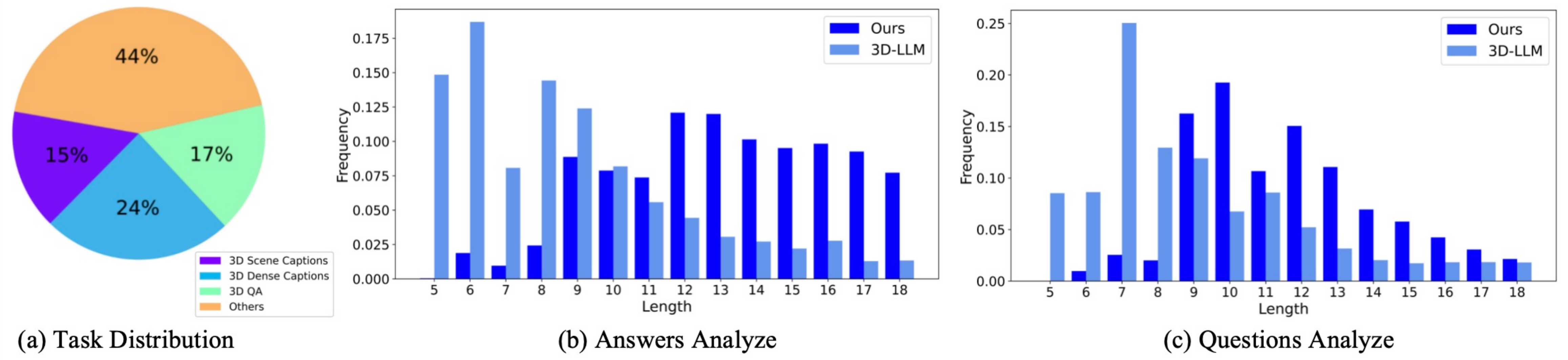}
\vspace{-1mm}
\caption{
(a) Task distribution across different categories. (b) Token frequency distribution in answers and (c) questions for the 3DS-160K dataset. The x-axis represents the number of tokens, and the y-axis denotes their frequency. Tokens are generated using the T5 tokenizer~\cite{raffel2020exploring}.
}
\vspace{-2mm}
\label{figure:distribution}
\end{figure*}

\subsection{Large Language Model}
Large Language Models (LLMs)\cite{devlin2018bert, raffel2020exploring, radford2018improving} have achieved significant advancements in recent years, driven by the availability of massive datasets and rapid progress in computational hardware. Models such as the encoder-only BERT\cite{devlin2018bert}, the encoder-decoder T5~\cite{raffel2020exploring}, and the decoder-centric GPT~\cite{radford2018improving} have demonstrated impressive performance on natural language processing (NLP) tasks, utilizing the Transformer architecture~\cite{vaswani2017attention}. The introduction of GPT-3~\cite{brown2020language} has further accelerated the adoption of decoder-only architectures, leveraging autoregressive decoding to produce coherent text predictions. Subsequent models, like PaLM~\cite{chowdhery2023palm}, have expanded the scale of model parameters and datasets, showcasing exceptional zero-shot learning capabilities. Moreover, models such as InstructGPT~\cite{ouyang2022training} have enhanced dialogue systems through instruction tuning, improving the quality of interactions. The field of NLP has also benefited from the contributions of open-source communities~\cite{chiang2023vicuna, touvron2023llama, yang2023baichuan}, which have facilitated collaborative and accessible approaches. However, as research increasingly ventures into the complexities of the physical world, the limitations of text-only modalities become more evident, highlighting the pressing need for LLMs to interpret and integrate data across diverse modalities.

\vspace{-0.5em}

\subsection{Multi-modal Large Language Model}
Researchers are actively exploring how to extend the capabilities of 
LLMs to address multi-modal problems, such as visual question answering and audio-visual representation. 
For instance, 
BLIP~\cite{li2022blip,li2023blip-2} series propose a Q-Former structure to facilitate cross-modal alignment between text and image modalities, 
bootstrapping vision-language pre-training from frozen pre-trained image encoders and frozen language decoders.
Flamingo~\cite{alayrac2022flamingo} integrates attention layers within the language model to fuse visual information. 
LLaVA~\cite{liu2023improved} utilizes simple linear layers for the projection of vision-language modalities and is tuned on an image-instruction-following dataset. 
Ferret~\cite{you2023ferret} further focused on finer visual granularity and extracts Region of Interest (ROI) features as visual tokens for region-level comprehension tasks.
Approaches like X-InstructBLIP~\cite{panagopoulou2023x} propose to expand language model capabilities to audio, video, and depth modalities, and OneLLM~\cite{han2023onellm} further seeks a unified encoder for all modality feature extraction. 
These developments demonstrate the potential of Multi-modal Large Language Models (MLLMs) for deeper understanding and richer interactions, pivotal for the seamless integration of diverse modalities by future AI systems. 
This warrants extensive research, especially given the potential applications in downstream tasks such as robot navigation, augmented reality, and virtual assistants, which could benefit from advancements in this domain.

\begin{figure*}[htbp]
\centering
\includegraphics[scale=0.23]{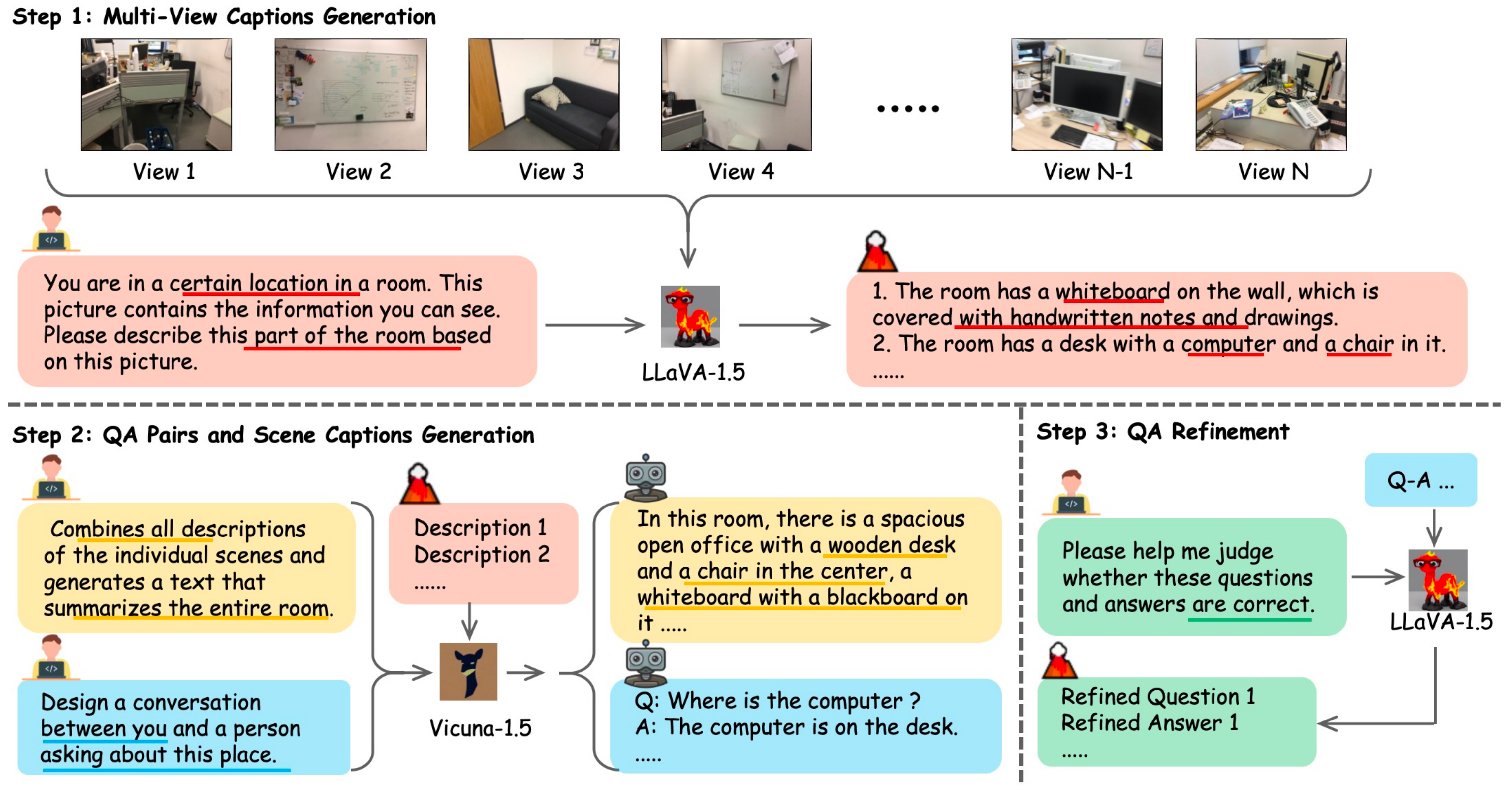}
\vspace{-1mm}
\caption{
The overview of the proposed data generation pipeline for 3DS-160K. The conceptual workflow mainly consists of three steps: 1) multi-view captions generation. 2) question-answer pairs and scene captions generation. 3) sample pairs refinement.
}
\vspace{-1mm}
\label{figure:3DS-160K}
\end{figure*}

\subsection{3D Scene Understanding}
3D scene understanding tasks require the model to interpret and respond to queries about the physical world perceptively using human language. This encompasses a suite of sub-tasks, each with distinct objectives: 3D Visual Question Answering (3DVQA)~\cite{azuma2022scanqa,ma2022sqa3d} challenges models to provide accurate answers to inquiries concerning the 3D environment; 
3D grounding~\cite{chen2020scanrefer} requires models to identify and localize regions within a scene as described by textual expressions; 
and 3D captioning~\cite{chen2021scan2cap} evaluates the capability to autonomously generate descriptive captions for 3D scenes. 
Numerous methods have been developed for fast and efficient perception of 3D spaces or objects. For instance, 3D object detection models~\cite{yin2022proposalcontrast, yin2021graph, yin2024fusion, yin2022proposalcontrast} offer precise object location information in point cloud space, providing valuable location priors for large language models.
Techniques like large sparse kernel training~\cite{feng2024lsk3dnet} enhance the efficiency of 3D environment perception. Shape2Scene (S2S)~\cite{feng2024shape2scene} learns 3D scene representations from shape data using MH-P/V backbones, demonstrating strong transferability in 3D tasks. A clustering-based supervised learning scheme~\cite{feng2023clustering} for point cloud analysis discovers latent subclass patterns by conducting within-class clustering on the point embedding space. Another approach~\cite{meng2020weakly} uses click-annotated BEV maps and a few precisely labeled instances in a two-stage architecture to generate cylindrical proposals and refine them to cuboids, achieving comparable performance with less data and serving as an annotation tool. Leveraging the comprehension and memorization capabilities of language models for 3D spatial-related tasks is an emerging research area. Models like PointLLM~\cite{xu2023pointllm} and ShapeLLM~\cite{qi2024shapellm} extract point cloud features with encoders such as ULIP~\cite{xue2023ulip}, OpenShape~\cite{liu2024openshape}, and ReCon~\cite{qi2023recon}, enhancing language models' understanding of object-level point clouds. Recently, 3D-LLM~\cite{hong20233d} employs multiple methods~\cite{hong20233d, kirillov2023segment, cheng2022masked} to reconstruct 3D spatial features and utilizes 2D Vision-Language model (VLM) to address understanding tasks within 3D scenes. Although the above approaches have made encouraging progress, in contrast, our 3UR-LLM adopts a end-to-end framework to efficiently parse the  3D point clouds and effectively stimulate MLLMs to understand complex physical world, facilitating a simplified reasoning pipeline while achieving superior performance with fewer training resources. 
\vspace{-1em}

\section{3DS-160K Benchmark}
The development of Large Language Models (LLMs) is significantly influenced by the availability of high-quality data~\cite{gao2023llama}. 
While acquiring 2D vision-language datasets is relatively straightforward through web scraping from abundant online resources, securing 3D-related data presents unique challenges due to the complexities and time demands of 3D data collection.
Currently, there are some 3D-text datasets, such as ScanQA~\cite{azuma2022scanqa} and SQA3D~\cite{ma2022sqa3d}, but they are designed with narrow focuses on particular 3D-related tasks. This restricts their wider application and limits their utility for downstream tasks, particularly in the context of embodied agents engaged in dynamic interactions within 3D environments.

3D-LLM~\cite{hong20233d} constructed a 3D-text dataset to pre-train MLLMs.
However, as illustrated in Figure~\ref{figure:distribution} (B) and (C), the dataset proposed by 3D-LLM exhibits an unhealthy distribution in the token length of the questions and answers. 
For instance, a significant portion of its responses are restricted to a narrow range of token counts, specifically from 0 to 7 tokens. 
This scarcity of tokens in the training data may hinder the model's ability to grasp the complexities and diversity of textual content~\cite{dong2023bamboo, bai2023longbench}. 
Therefore, we deliver to establish a data generation pipeline that aims to refine the distribution of response lengths and equip the model with the capability to accommodate a wider array of textual diversity.

\subsection{3D-text Data Generation Pipeline}
We devise a 3D-text data generation pipeline based on open-source 2D MLLM~\cite{liu2024visual} and LLM~\cite{chiang2023vicuna}, for generating high-quality 3D-text pairs.
As illustrated in Fig.~\ref{figure:3DS-160K}, the detailed generation steps for paired textual data mainly involve the following phases:
\begin{itemize} [leftmargin=0cm, itemindent=0.5cm]
    \item[i)]{
    \textit{Multi-view Captions Generation}. 
    We initiate with periodic sampling of 2D images from point cloud data to obtain a discrete representation of the 3D scene. These varied-perspective images are input into LLaVA~\cite{liu2024visual} to generate descriptive narratives in response to specific prompt languages, yielding multiple scene interpretations that reflect diverse spatial orientations and encompass both detailed and holistic scene aspects. To mitigate the risk of producing descriptions with hallucinated content, we employ a controlled generation temperature setting.
    }
    \item[ii)]{
    \textit{
    QA Pairs and Scene Captions Generation.
    }
    Utilizing the multi-view narratives, we engage Vicuna~\cite{chiang2023vicuna} to leverage its advanced in-context learning abilities for the creation of multi-turn dialogues and comprehensive 
    captions that incorporate a global viewpoint.
    }
    \item[iii)]{
    \textit{QA Refinement.}
    For quality assurance of the outputs, we reintroduce the multi-modal samples into LLaVA~\cite{liu2024visual} for refinement. This process entails LLaVA reassessing the dialogues alongside their associated imagery, directed by image indices, to guarantee the textual and visual data congruence.
    }
\end{itemize}

\subsection{Construction of 3DS-160K}

Our 3D-text pairs are derived from two prominent 3D scene datasets: ScanNet~\cite{dai2017scannet} and 3RScan~\cite{wald2019rio}. ScanNet~\cite{dai2017scannet} includes 1.5K richly annotated 3D indoor scenes, featuring multi-view images with segmentation maps and bounding boxes. 3RScan~\cite{wald2019rio} provides a collection of 1.4K 3D reconstructions from 478 dynamically evolving indoor environments. Through the carefully crafted prompts and a rigorous generation and refinement process, we have created 160K high-quality 3D-text samples, compiled into the 3DS-160K dataset. The 3DS-160K dataset supports three primary tasks: 3D dense captioning, 3D visual question answering  and 3D scene captioning, and includes a total of 163,269 3D-text pairs.

This multi-task framework is designed to help models acquire cross-task representational patterns~\cite{chen2023minigpt}, thereby enhancing their adaptability and robustness across a wide range of downstream applications~\cite{radford2019language}. The distribution of textual tokens in 3DS-160K follows a natural trend. As illustrated in Fig.~\ref{figure:distribution}, the distribution of question lengths is more balanced, with 40.8\% of questions containing fewer than 16 tokens, compared to 73.3\% in the benchmark 3D-LLM dataset. Additionally, 3DS-160K includes a higher proportion of answers exceeding 7 tokens (68.1\% versus 13.9\% in 3D-LLM), encouraging models to achieve a more precise alignment between 3D scenes and textual descriptions. We believe that the 3DS-160K dataset will significantly enrich the resources available to the 3D scene understanding community.

\begin{figure*}[htbp]
\centering
\setlength{\abovecaptionskip}{0.2cm}
\includegraphics[scale=0.22]{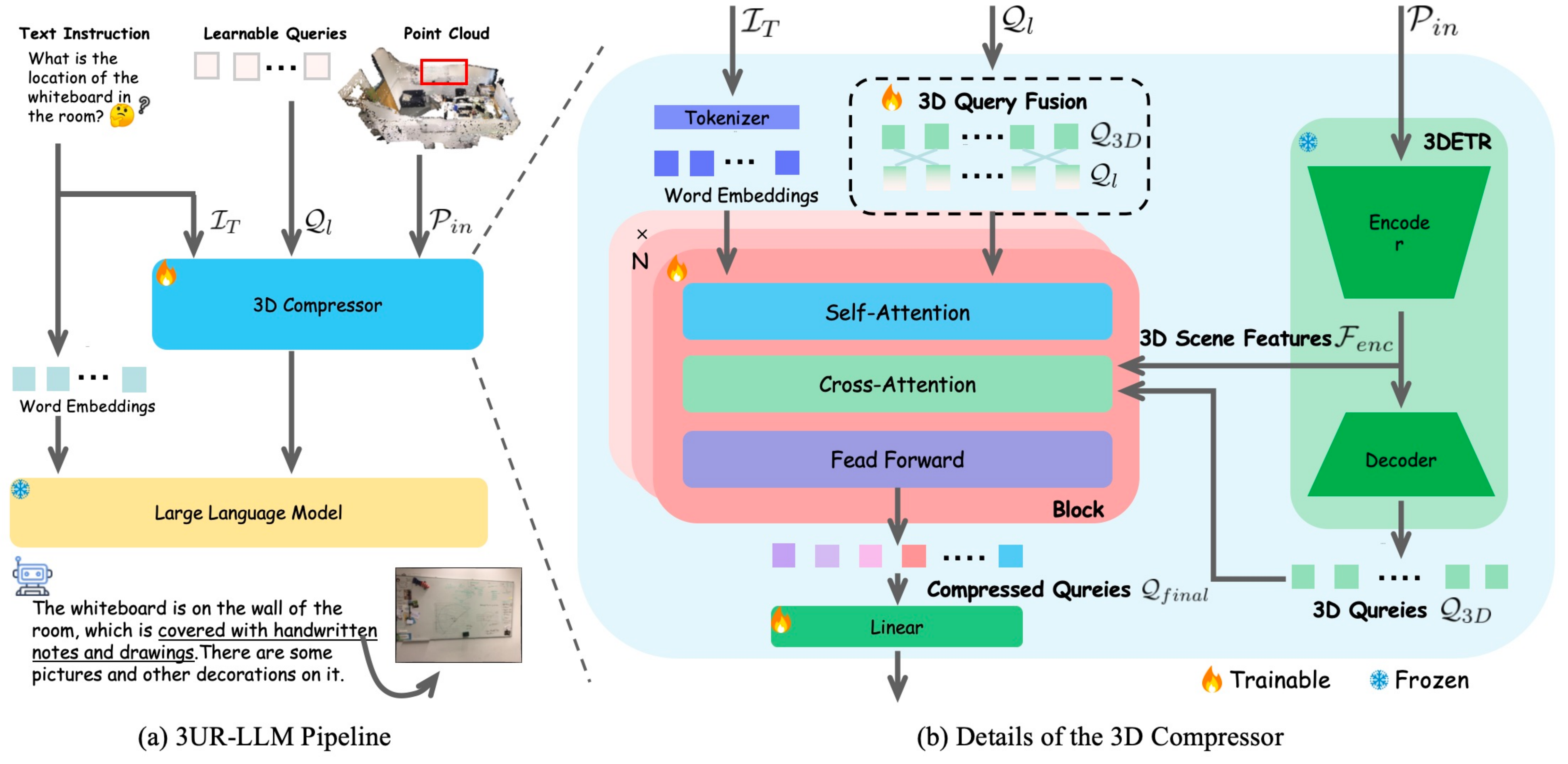}
\vspace{-2mm}
\caption{The pipeline of our 3UR-LLM framework.
(a) Overview of the pipeline. The 3D compressor consolidates scene features and textual tokens into unified queries, which are then fed into a large language model to generate textual responses.
(b) Details of the 3D compressor which condensed 3D features and textual tokens into compact queries.} 
\vspace{-2mm}
\label{figure:4}
\end{figure*}




\section{3UR-LLM Framework}
In this section, we provide a detailed overview of the 3UR-LLM framework. We begin by discussing the spatial encoder in 3UR-LLM, which is responsible for 3D vision perception. Given the limited availability of large-scale pre-trained modality encoders like CLIP~\cite{radford2021learning}, we employ an encoder that leverages knowledge from a well-trained 3D object detector~\cite{misra2021end}. Next, we describe the 3D compressor structure, which is designed to efficiently map and refine 3D features into a manageable set of tokens, accompanied by corresponding text descriptions. Finally, we introduce the 3D query fusion mechanism, which further enhances and integrates high-confidence queries for improved spatial understanding and optimized computational efficiency. The overall framework of 3UR-LLM is illustrated in Fig.~\ref{figure:4}.

\subsection{Spatial Encoder}
The 3D spatial encoder is the pillar of vision perception in 3D scene understanding.
Traditional approaches, such as those employing 3D Convolutions~\cite{choy20194d}, are often limited to structured inputs like voxels, which reduces their effectiveness when handling unstructured data, such as point clouds. 
In contrast, Transformer architectures~\cite{zhao2021point} demonstrate superior capabilities in capturing global dependencies and are robust to data permutation, making them highly promising for processing complex 3D spatial structures. Despite these advancements, maintaining the compactness of the 3D perception component is crucial.
While 3D-LLM~\cite{hong20233d} uses advanced techniques~\cite{hong20233d,jatavallabhula2020slam} to reconstruct 3D scenes from multi-view images, the complexity of these processes can impede efficient extraction of point cloud information. To address this challenge, we incorporate a transformer-based 3D object detector, 3DETR~\cite{misra2021end}, as the foundation of our 3D vision perception module. This approach enables the capture of diverse visual cues, which are then integrated into the subsequent language model, including global features $\mathcal{F}{enc} \in \mathbb{R}^{N{enc} \times C}$ from the encoder and hierarchical features $\mathcal{Q}{3D} \in \mathbb{R}^{N{3D} \times C}$ from the decoder.
The 3D object perception process is encapsulated by:
\begin{equation}
    \mathcal{F}_{enc} = Encoder(\mathcal{P}_{in}),
\end{equation}
\begin{equation}
    \mathcal{Q}_{3D} = Decoder(\mathcal{F}_{enc}, \mathcal{Q}_{3D}),
\end{equation}
Compared to 3D-LLM~\cite{hong20233d}, 3UR-LLM efficiently extracts features directly from point clouds, leveraging prior knowledge from the detection model and providing object-level perception for model training. Our experiments demonstrate that this streamlined approach significantly enhances 3D understanding accuracy and offers considerable efficiency gains, as discussed in Section~\ref{sec:ablation}.

Nonetheless, the feature extraction process presents several challenges, particularly the excessive length of feature sequences. For example, when processing a single point cloud scene, we randomly sample 4K points, which are then encoded by 3DETR~\cite{misra2021end} to produce $\mathcal{F}{enc} \in \mathbb{R}^{1024 \times C}$ and $\mathcal{Q}{3D} \in \mathbb{R}^{256 \times C}$. Converting these feature sequences entirely into tokens becomes impractical due to the exponential increase in memory consumption imposed by the self-attention mechanism~\cite{vaswani2017attention}. Additionally, aligning these features with textual features is essential for the language model to accurately interpret spatial semantic information~\cite{zhu2023languagebind, liu2023improved, li2023blip-2}.
To address these challenges, we propose a methodology that compresses the 3D spatial features into a manageable set of tokens containing essential information while effectively aligning them with the textual embeddings.


\subsection{3D Compressor}
We adopt a multi-layer transformer architecture as the foundation for designing our 3D compressor. This module leverages a series of learnable queries within $\mathcal{N}$ stacked transformer blocks, enabling effective refinement and management of feature sequence lengths. Specifically, we initialize the learnable queries as $\mathcal{Q}_l \in \mathbb{R}^{N_q \times C_q}$, which encapsulates compressed spatial information. 
For the text instructions ${\mathcal{I}_T}$, we use a tokenizer to convert them into word embeddings $\mathcal{F}_T \in \mathbb{R}^{N_t \times C_t}$, which are then concatenated with the initialized queries to facilitate self-attention computations. To ensure alignment between the compressed features and the textual embeddings, we employ a linear layer to adjust the dimensionality of the queries from $C_q$ to $C_t$.
The compression process can be represented by the following formula:
\begin{equation}
    \mathcal{F}_{c} = \mathcal{C}(\mathcal{P}(\mathcal{Q}_l), \mathcal{F}_T),
\end{equation}
\begin{equation}
    \mathcal{F}_{s} = softmax\left(\frac{\mathcal{F}_{c}\mathcal{F}_{c}^T}{\sqrt{d_k}}\right)\mathcal{F}_{c},
\end{equation}
where $\mathcal{P}$ and $\mathcal{C}$ represent linear projection and concatenation in sequence dimension, $d_k$ is the dimension of self-attention layer.

Afterwards, cross-attention mechanisms are employed to facilitate cross-modality information interaction. For the visual features $\mathcal{F}_{enc}$ and $\mathcal{Q}_{3D}$ 
, we employ a linear layer  to transform their dimensions and concat them to form $\mathcal{K}, \mathcal{V} \in R^{(N_{enc} + N_{3D}) \times C}$. 
From this concatenated set, we extract the queries $\mathcal{Q} \in \mathbb{R}^{N_q \times C}$, enabling them to directly interact with the visual features as:
\begin{equation}
    \mathcal{Q} = \mathcal{S}(\mathcal{F}_{s}), 
\end{equation}
\begin{equation}
    \mathcal{K=V} = \mathcal{C}(\mathcal{P}(\mathcal{F}_{enc}), \mathcal{P}(\mathcal{Q}_{3D})), 
\end{equation}
\begin{equation}
    \mathcal{F}_{final} = FFN(softmax\left(\frac{\mathcal{Q}\mathcal{K}^T}{\sqrt{d_k}}\right)\mathcal{V}),
\end{equation}
where $\mathcal{S}$ denotes the selection of the first $N_q$ tokens from $\mathcal{F}{s}$, and $FFN$ refers to a feed-forward network.
Finally, the condensed visual features $\mathcal{F}_{final} \in \mathbb{R}^{N_q \times C}$, generated by the 3D compressor, are projected into the linguistic embedding space to facilitate response generation.
\begin{align}
    \mathcal{Q}_{final} = \mathcal{P}(\mathcal{F}_{final}) \in R^{N_q \times C},
\end{align}
\subsection{3D Query Fusion}
We devise a 3D query fusion module to equip $\mathcal{Q}_{l}$ with a better perception of objects in space during the information compression.
Specifically, it uses the objectness probability $\mathcal{P}_{obj}$ output from $\mathcal{Q}_{3D}$ as a criterion, which is mainly used to determine whether the current query contains certain object information.
It first selects the top $N_q$ from $\mathcal{Q}_{3D}$ and aligns their feature dimensions with $\mathcal{Q}_{l}$ through a linear layer. 
Finally, it adds them up
to update the parameters of $\mathcal{Q}_{l}$.
The entire computation process is represented by:
\begin{equation}
    idx = top_k(\mathcal{P}_{obj}), 
\end{equation}
\begin{equation}   
    \mathcal{Q}_{3D}^{'} = \mathcal{P}(\mathcal{Q}_{3D}^{idx}),
\end{equation}
\begin{equation} 
    \mathcal{Q}_{l}^{'} = \mathcal{Q}_{l} + \mathcal{Q}_{3D}^{'}, 
\end{equation} 
where we use the updated $\mathcal{Q}_{l}^{'}$ as the newly initialized queries for feature compression.

\begin{figure}[t]
\centering
\setlength{\abovecaptionskip}{0.cm}
\includegraphics[scale=0.24]{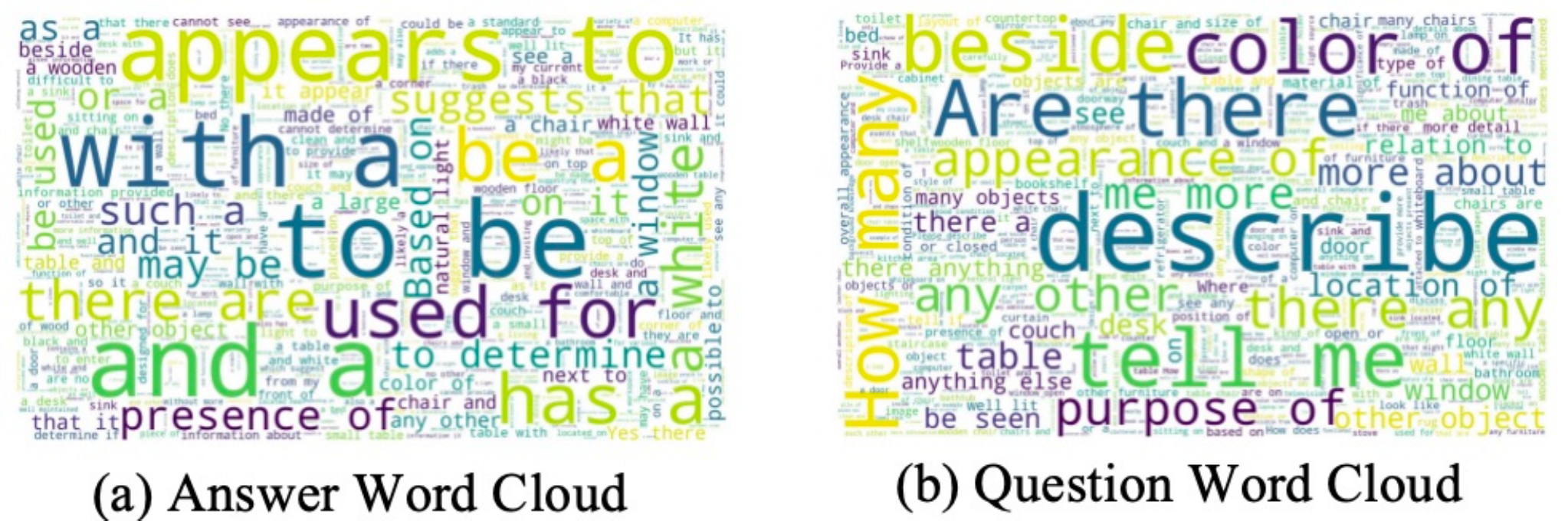}
\vspace{-2mm}
\caption{
Word clouds depicting the distribution of words in answers (a) and questions (b) within our 3DS-160K dataset.
}
\vspace{-2mm}
\label{figure:word_cloud}
\end{figure}



\subsection{Loss Function}
The goal of our training process is to optimize the model parameters $\Theta$ to maximize the likelihood of generating the target response sequence $\mathcal{S}$, given the input point cloud $\mathcal{P}_{in}$ and the text interactions $\mathcal{I}_T$. This optimization objective can be expressed as:
\begin{equation}
    \Theta^{*} = arg max(\mathcal{S}|\mathcal{P}_{in};\mathcal{I}_T;\Theta),
\end{equation}
In practice, we achieve this by employing a token-wise cross-entropy loss, which trains the model to predict the $i$-th token $\mathcal{S}{[i]}$ based on the preceding $i-1$ tokens $\mathcal{S}{[1,...,i-1]}$. The loss function is formulated as:
\begin{equation}
   \mathcal{L}(\Theta) = - \sum_{i=1}^{|\mathcal{S}|}log P(\mathcal{S}_{[i]} | \mathcal{P}_{in};\mathcal{I}_T;\Theta;\mathcal{S}_{[1,...,i-1]}), 
\end{equation}
where $|\mathcal{S}|$ is the number of tokens in the desired response.

\section{Experiment}

\begin{table*}[htb]   
\renewcommand{\arraystretch}{1.3}
\tabcolsep=0.20cm
\caption{Quantitative comparison of model performance on ScanQA. “3D” refers to whether the model receives point clouds as input. 
We categorize all methods into classification (CLS) and generation (GEN) methods. The results of 3D-LLM comes from our fine-tuned version and using Flan-T5-XL as LLM backbone.}
\vspace{-2mm}
\begin{center}   
\begin{tabular}{c | c | c | c c c c c c c c}   
\hline   
{Method} & {3D} & {Type} & {BLEU-1} & {BLEU-2} & {BLEU-3} & {BLEU-4} & {METEOR} & {ROUGE-L} & {CIDEr} & {EM@1} \\
\hline   
ScanQA\cite{azuma2022scanqa}         & $\checkmark$ & \multirow{2}*{CLS} & 30.2 & 20.4 & 15.1 & 10.1 & 13.1 & 33.3 & 64.9 & 21.0 \\
3D-VLP\cite{jin2023context}         & $\checkmark$ &             ~               & 30.5 & 21.3 & \underline{16.6} & 11.1 & 13.5 & 34.5 & 66.9 & 21.6 \\ \hline
VoteNet+MCAN\cite{hong20233d}   & $\checkmark$ & \multirow{5}*{GEN} & 28.0 & 16.7 & 10.8 & 6.2 & 11.4 & 29.8 & 54.7 & 17.3 \\
ScanRefer+MCAN\cite{hong20233d} & $\checkmark$ &             ~               & 26.9 & 16.6 & 11.6 & 7.9 & 11.5 & 30.0 & 55.4 & 18.6 \\
3D-LLM\cite{hong20233d}         & $\times$     &             ~               & {39.3} & {25.2} & \underline{18.4} & {12.0} & {14.5} & {35.7} & {69.4} & {20.5} \\
LL3DA\cite{chen2024ll3da}   & $\checkmark$     &             ~              & \underline{41.2} & \underline{26.7} & \underline{19.2} & \underline{13.5} & {15.8} & {37.3} & {76.7} & {20.9} \\
3UR-LLM & $\checkmark$ &   ~   & {39.7} & {25.8} & {18.4} & {12.7} & {15.8} & {38.2} & {76.5} & \underline{21.3} \\
\textcolor{red}{{3UR-LLM-mix}} & \textcolor{red}{$\checkmark$} &   ~   & \textcolor{red}{40.0} & \textcolor{red}{26.1} & \textcolor{red}{18.5} & \textcolor{red}{12.4} & \textcolor{red}{\underline{15.9}} & \textcolor{red}{\textbf{38.4}} & \textcolor{red}{\underline{76.9}} & \textcolor{red}{20.6} \\
\textcolor{red}{{3UR-LLM+}} & \textcolor{red}{$\checkmark$} &   ~   & \textcolor{red}{\textbf{43.7}} & \textcolor{red}{\textbf{30.1}} & \textcolor{red}{\textbf{22.1}} & \textcolor{red}{\textbf{15.5}} & \textcolor{red}{\textbf{18.4}} & \textcolor{red}{\textbf{41.5}} & \textcolor{red}{\textbf{87.7}} & \textcolor{red}{\textbf{21.5}} \\
\hline   
\end{tabular}   
\vspace{-2mm}
\label{table:1} 
\end{center}   
\end{table*}

\subsection{Datasets and Metrics}
  
We pre-train the 3UR-LLM on 3DS-160K, which includes point cloud data from the training set of ScanNet~\cite{dai2017scannet} and 3RScan~\cite{wald2019rio}. The composition of 3DS-160K is visually represented through a word cloud in Table~\ref{figure:word_cloud}.
To validate the model's versatility across various 3D understanding tasks, we fine-tune 3UR-LLM on the ScanQA~\cite{azuma2022scanqa} and SQA3D~\cite{ma2022sqa3d} benchmarks. The model's performance is evaluated using a suite of metrics, including CIDEr~\cite{vedantam2015cider}, BLEU~\cite{papineni2002bleu}, METEOR~\cite{banerjee2005meteor}, and ROUGE-L~\cite{lin2004rouge}, to assess the quality of generated text responses.
\textcolor{red}{In addition, we introduce an enhanced variant of our model, 3UR-LLM+, which incorporates the more advanced Mask3D~\cite{schult2023mask3d} as the 3D encoder and adopts vicuna-7B-v1.5~\cite{zheng2023judging} as the upgraded LLM backbone.}

\begin{figure*}
    \begin{minipage}{1.0\linewidth}
		\vspace{1pt}
		\centerline{\includegraphics[scale=0.235]{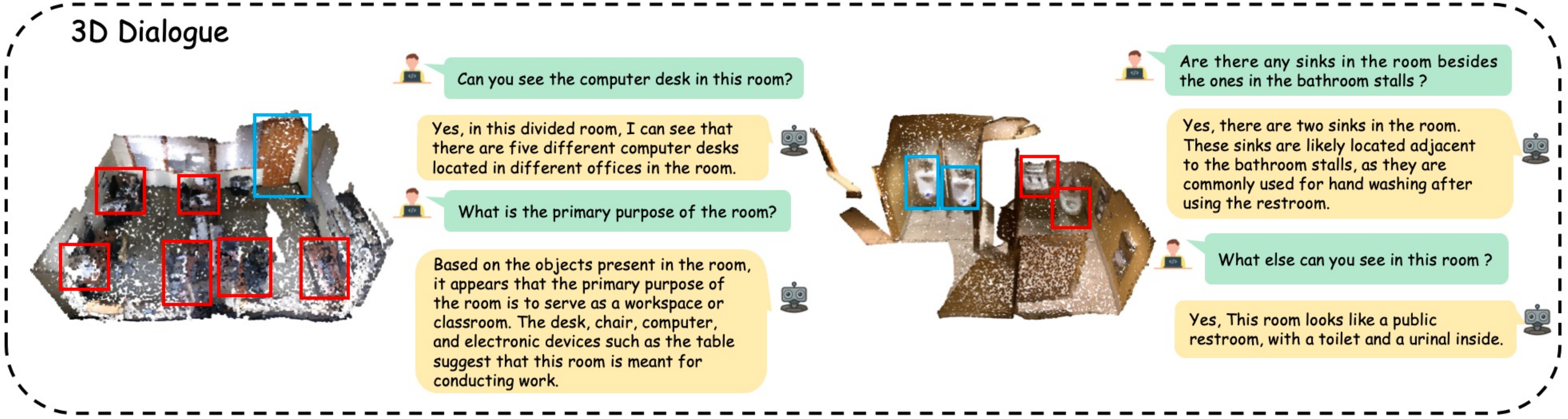}}
			\vspace{1pt}
		\centerline{\includegraphics[scale=0.235]{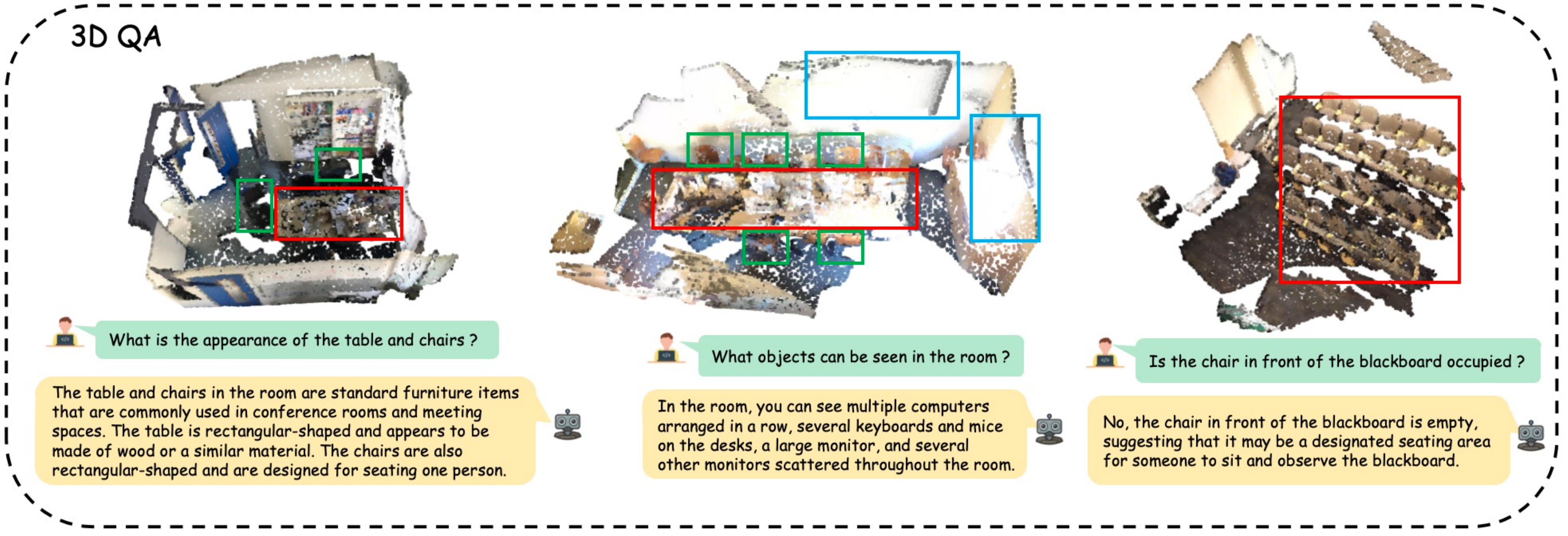}}
			\vspace{1pt}
		\centerline{\includegraphics[scale=0.235]{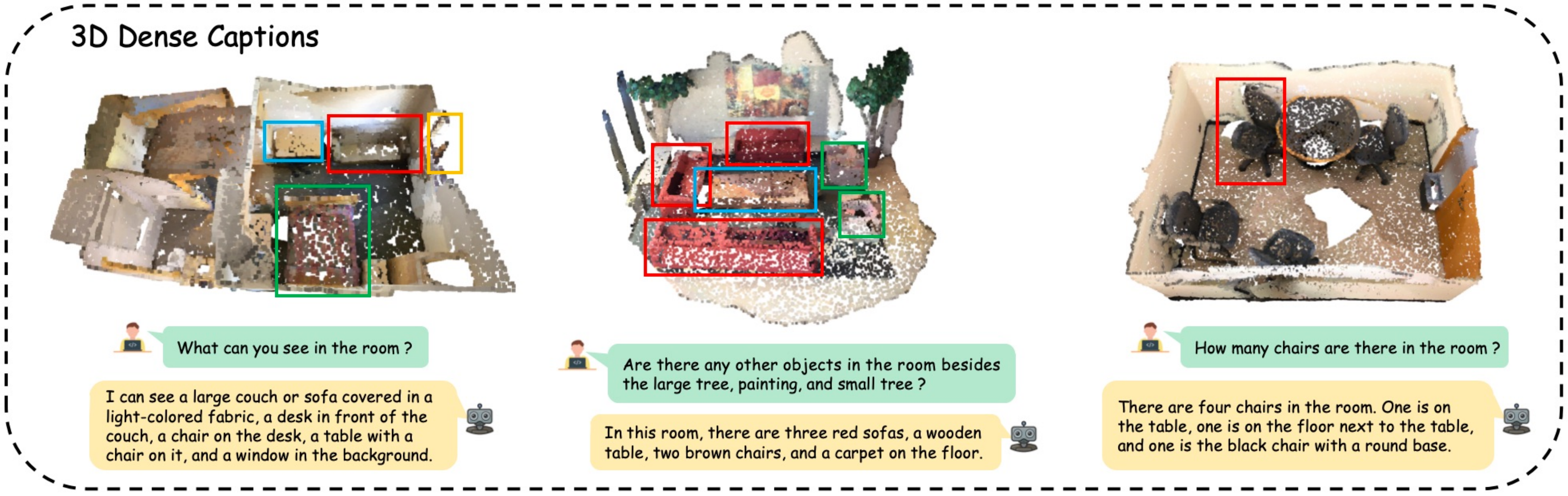}}
        \centerline{\includegraphics[scale=0.235]{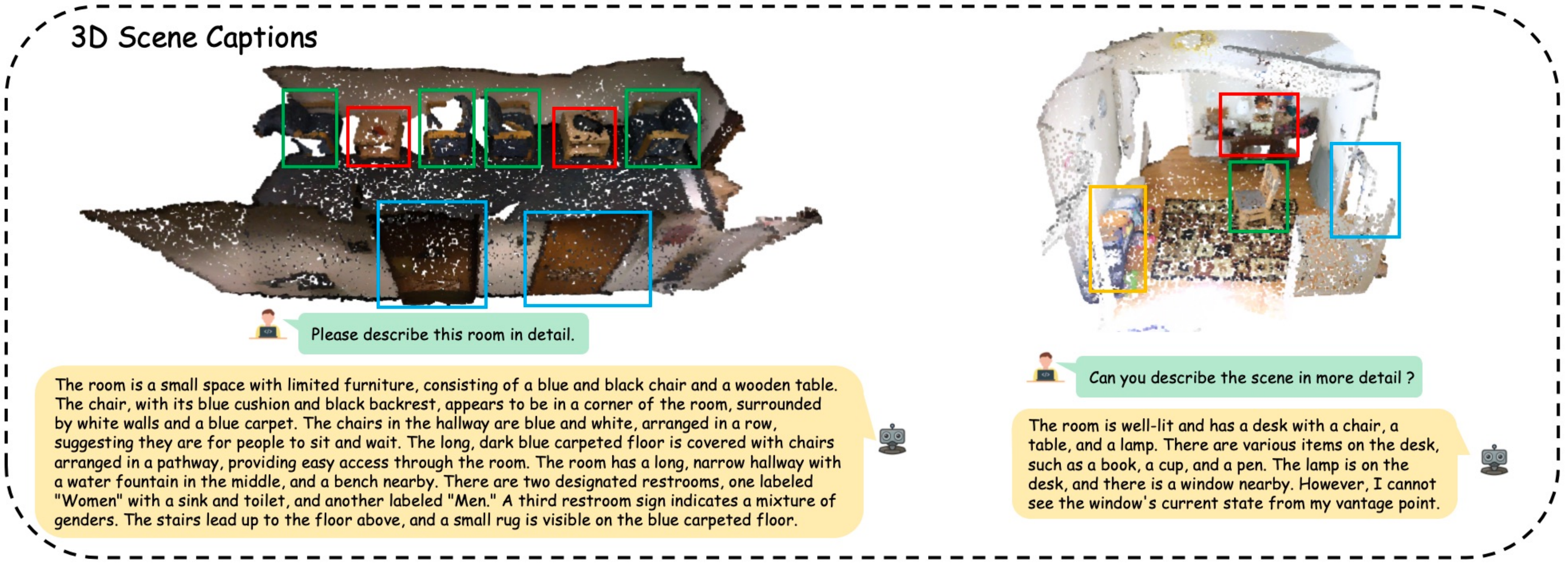}}
	\end{minipage}
    \caption{
 Representative visualizations from the 3DS-160K dataset, showcasing a range of perception tasks including 3D dialogue, dense captioning, scene captioning, and question answering. The diversity of these data enhances the model's capability in handling various downstream tasks.}
    \label{figure:cases_in_3ds}
\end{figure*}

Fig.~\ref{figure:cases_in_3ds} provides a detailed visualization of samples from 3DS-160K, illustrating the dataset's breadth across common 3D tasks such as 3D Dialogue, 3D Question Answering (QA), 3D Dense Captions, and 3D Scene Captions. The dataset exhibits notable advantages over existing 3D-text pre-training datasets~\cite{hong20233d, zhu20233d}, such as enhanced multi-object interactivity, multi-round dialogue capabilities, and a greater diversity of 3D scenes.

\subsection{Implementation Details}
To ensure a fair comparison, we use Flan-T5-XL~\cite{chung2022scaling} as our LLM backbone, aligning with the architecture employed in 3D-LLM~\cite{hong20233d}. To minimize memory usage, the language model remains frozen during both the pre-training and fine-tuning phases. Consistent with prior work in 3D visual tasks~\cite{misra2021end, chen2021scan2cap}, we sample 4K point clouds from each scene as our 3D visual input.

During the pre-training phase, we train 3UR-LLM for 20 epochs on 4 Nvidia A800 GPUs, with a total batch size of 16. We use AdamW~\cite{loshchilov2017decoupled} as the optimizer, applying a weight decay of 0.1, and a cosine annealing scheduler to adjust the learning rate from $10^{-4}$ to $10^{-5}$. For the instruction tuning phase, the learning rate is configured to decrease from $10^{-4}$ to $10^{-6}$, with the same optimizer. In this phase, we fine-tune 3UR-LLM on 2 Nvidia A800 GPUs for a total of 100 epochs.


\begin{figure*}[htbp]
\centering
\setlength{\abovecaptionskip}{0.cm}
\includegraphics[scale=0.21]{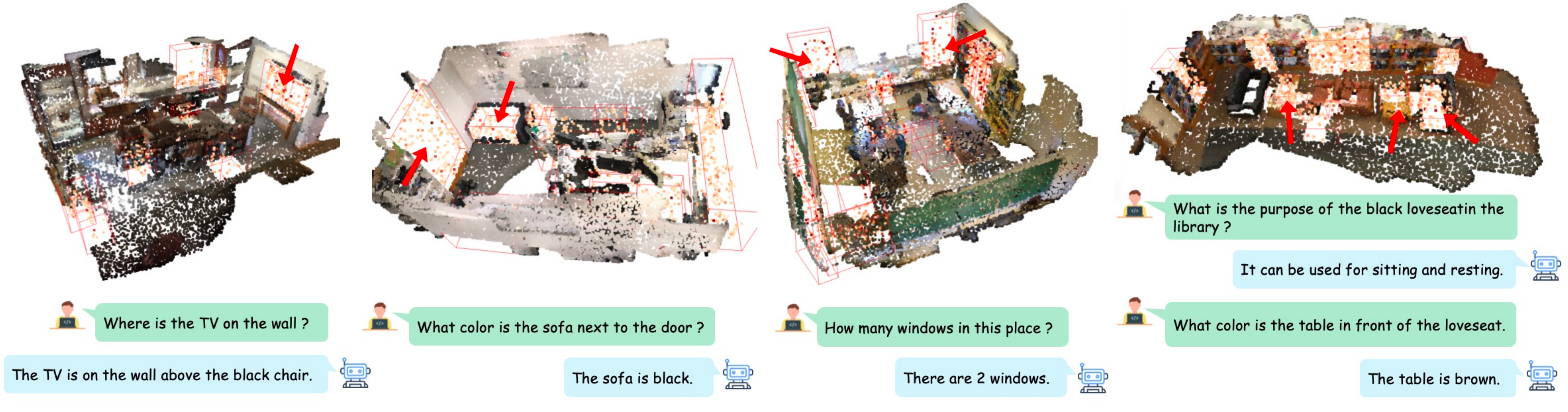}
\vspace{-2mm}
\caption{
Visualization 
on object-level feature perception and inter-object relationship understanding.
The top-10 queries are projected back onto the original point cloud, providing an intuitive representation of object-level features, with red points highlighting areas of higher response. 3UR-LLM precisely associates objects' spatial positions with their textual descriptions, demonstrating robust understanding of complex 3D scenes involving diverse objects.
}
\vspace{-2mm}
\label{figure:attention}
\end{figure*}

\subsection{Result Analysis}
Table~\ref{table:1} and \ref{table:2} present a comparative analysis of our proposed 3UR-LLM against existing methods on the ScanQA\cite{azuma2022scanqa} and SQA3D~\cite{ma2022sqa3d} benchmarks. Methods are categorized into two groups: Classification Models (CLS), which select answers from predefined options, including ScanQA~\cite{azuma2022scanqa}, 3D-VLP~\cite{jin2023context}, 3D-Vista~\cite{zhu20233d}, and ClipBERT~\cite{ma2022sqa3d}; and Text Generation Models (GEN), which produce response sentences. Notable generation models include ScanRefer+MCAN~\cite{hong20233d}, VoteNet+MCAN~\cite{hong20233d}, and 3D-LLM~\cite{hong20233d}, where ScanRefer+MCAN and VoteNet+MCAN combine object identification via ScanRefer~\cite{chen2020scanrefer} and VoteNet~\cite{ding2019votenet} with text generation through MCAN~\cite{yu2019mcan}.

Our results highlight that 3UR-LLM not only surpasses classification models on the ScanQA validation set but also exceeds the performance of 3D-LLM, achieving a +7.1\% increase in CIDEr and a +2.5\% increase in ROUGE-L scores. Similarly, on the SQA3D benchmarks, 3UR-LLM demonstrates superior performance, outperforming both classification and generation models with an average accuracy improvement of +1.5\% and +1.9\%, respectively.
\textcolor{red}{Equipped with a more robust 3D encoder and LLM, our enhanced model, 3UR-LLM+, demonstrates superior performance over LL3DA, achieving improvements of 2.0\% in BLEU-1, 11.0\% in CIDEr, and 4.2\% in ROUGE-L on the ScanQA benchmark. Additionally, on SQA3D, 3UR-LLM+ not only improves average accuracy by 3.7\% compared to its predecessor, 3UR-LLM, but also surpasses LL3DA by 1.2\% in average accuracy.}

\begin{table}[t]   
\renewcommand{\arraystretch}{1.3}
\caption{Quantitative performance comparison on SQA3D.}  
\vspace{-2mm}
\tabcolsep=0.05cm
\begin{center}   
\begin{tabular}{c | c | c | c c c c c c c }   
\hline   
{Method} & {3D} & {Type} & {What} & {Is} & {How} & {Can} & {Which} & {Others} & {Avg} \\
\hline   
SQA3D\cite{ma2022sqa3d}                      & $\checkmark$ & \multirow{3}*{CLS} & 31.6 & 63.8 & \textbf{46.0} & \underline{69.5} & 43.0 & 46.4 & 47.2  \\
3D-Vista\cite{zhu20233d} & $\checkmark$ &             ~               & 34.8 & 63.3 & 45.4 & \textbf{69.8} & 47.2 & \underline{48.1} & 48.5  \\ 
ClipBERT\cite{ma2022sqa3d} & $\times$     &             ~               & 30.2 & 60.1 & 38.7 & 63.3 & 42.5 & 42.7 & 43.3  \\ \hline
3D-LLM\cite{hong20233d}                     & $\times$     & \multirow{4}*{GEN} & 35.5 & \textbf{64.7} & 43.8 & 68.6 & 48.7 & 45.2 & 48.1  \\ 
\textcolor{red}{LL3DA~\cite{chen2024ll3da}} & \textcolor{red}{$\checkmark$} &             ~               & \textcolor{red}{\underline{42.7}} & \textcolor{red}{\underline{64.1}} & \textcolor{red}{43.9} & \textcolor{red}{67.7} & \textcolor{red}{\underline{49.2}} & \textcolor{red}{\underline{52.7}} & \textcolor{red}{\underline{52.5}}  \\ 
3UR-LLM                     & $\checkmark$ &             ~               & 38.0 & 63.8 & \underline{45.8} & 68.6 & 47.8 & 51.6 & 50.0 \\
\textcolor{red}{3UR-LLM+}  & \textcolor{red}{$\checkmark$} &             ~               & \textcolor{red}{\textbf{48.6}} & \textcolor{red}{\textbf{64.7}} & \textcolor{red}{44.9} & \textcolor{red}{68.3} & \textcolor{red}{\textbf{54.3}} & \textcolor{red}{\textbf{52.9}} & \textcolor{red}{\textbf{53.7}} \\
\hline   
\end{tabular}   
\vspace{-4mm}
\label{table:2} 
\end{center}   
\end{table}

Moreover, we visualize some representative cases in Fig.~\ref{figure:attention} to confirm the effectiveness of object-level 3D feature perception.
The features $\mathcal{Q}_{3D}$ provided by 3DETR~\cite{misra2021end} exhibit a strong correlation with the object information in the text, thereby further promoting 3UR-LLM to better comprehend the cross-spatial object relationships. 
The features remain stable in the face of diverse questions.
\vspace{-0.5em}

\subsection{Efficiency Analysis}

To evaluate model efficiency, we focus on two key aspects: feature extraction and training. For feature extraction, we measure the time required to project a scene-level point cloud from the 3D compressor to the language model, providing insight into the feature extraction efficiency. In the case of 3D-LLM~\cite{hong20233d}, which involves three distinct steps for feature extraction, we calculate the total time for these steps, including the time consumed by the perceiver module.
Our 3UR-LLM integrates 3D feature extraction and processing into a single end-to-end framework, which results in a substantial reduction in overall time, saving 187 GPU hours. For training efficiency, we compare the resource consumption in terms of training time and GPU memory usage. Training time is defined as the total GPU hours required for both the pre-training and fine-tuning phases. As shown in Table~\ref{table:time and memory}, while 3UR-LLM incorporates a 3D compressor with more learnable parameters, leading to a 25\% increase in GPU memory usage during training compared to 3D-LLM~\cite{hong20233d}, it also achieves a 49\% reduction in time cost.

\begin{table}[t]   
\caption{
Analysis of training time and memory usage.
Training time is measured in GPU hours, and memory usage is reported in gigabytes (GB). "FE" refers to additional feature extraction.
} 
\renewcommand{\arraystretch}{1.2}
\tabcolsep=0.23cm
\begin{center}   
\begin{tabular}{c | c  c  c  c  c}   
\hline   
\multirow{2}*{Method} & \multicolumn{4}{c}{time cost} & \multirow{2}*{memory} \\
\cline{2-5}
~ & {FE} & {pre-train} & {fine-tune} & {total} & ~ \\
\hline   3D-LLM & 266 & 64 & 45 & 379 & 28   \\
Ours   &  $\times$  & 96 & 96 & 192 & 35  \\ 
\hline   
\end{tabular} 
\vspace{1.5mm}
\label{table:time and memory} 
\vspace{-3em}
\end{center}  
\end{table}

\begin{table*}[htb]   
\renewcommand{\arraystretch}{1.2}
\caption{
Ablation study on the 3D query fusion module.
“Trainable” item means whether keep the 3D compressor frozen during fine-tuning, and “3D-QF” represents whether use 3D query fusion.
}
\tabcolsep=0.08cm
\begin{center}   
\begin{tabular}{c c | c c c c c c c c}   
\hline   

{Trainable} & {3D-QF} & {BLEU-1} & {BLEU-2} & {BLEU-3} & {BLEU-4} & {METEOR} & {ROUGE-L} & {CIDEr} & {EM@1}\\
\hline
~ & ~  & 38.03\scriptsize{($\textcolor[RGB]{61,154,64}{ -1.71 \%}$)} & 23.25\scriptsize{($\textcolor[RGB]{61,154,64}{ -2.57 \%}$)} & 16.39\scriptsize{($\textcolor[RGB]{61,154,64}{ -2.04 \%}$)} & 11.26\scriptsize{($\textcolor[RGB]{61,154,64} {-1.46 \%}$)} & 14.98\scriptsize{($\textcolor[RGB]{61,154,64} {-0.85 \%}$)} & 37.74\scriptsize{($\textcolor[RGB]{61,154,64} {-0.50 \%}$)} & 73.87\scriptsize{($\textcolor[RGB]{61,154,64} {-2.64 \%}$)} & 20.16\scriptsize{($\textcolor[RGB]{61,154,64} {-1.18 \%}$)}  \\ 
$\checkmark$    & ~  & 38.24\scriptsize{($\textcolor[RGB]{61,154,64} {-1.50 \%}$)} & 24.14\scriptsize{($\textcolor[RGB]{61,154,64} {-1.68 \%}$)} & 16.52\scriptsize{($\textcolor[RGB]{61,154,64} {-1.91 \%}$)} & 11.55\scriptsize{($\textcolor[RGB]{61,154,64} {-1.17 \%}$)} & 15.70\scriptsize{($\textcolor[RGB]{61,154,64} {-0.13 \%}$)} & 37.84\scriptsize{($\textcolor[RGB]{61,154,64} {-0.40 \%}$)}  & 74.16\scriptsize{($\textcolor[RGB]{61,154,64} {-2.35 \%}$)} & 20.43\scriptsize{($\textcolor[RGB]{61,154,64} {-0.91 \%}$)}   \\
~ & $\checkmark$ & \underline{38.56}\scriptsize{($\textcolor[RGB]{61,154,64} {-1.18 \%}$)} & \underline{24.95}\scriptsize{($\textcolor[RGB]{61,154,64} {-0.87 \%}$)} & \underline{17.63}\scriptsize{($\textcolor[RGB]{61,154,64} {-0.80 \%}$)} & \underline{12.23}\scriptsize{($\textcolor[RGB]{61,154,64} {-0.49 \%}$)} & \underline{15.74}\scriptsize{($\textcolor[RGB]{61,154,64} {-0.09 \%}$)} & \underline{38.15}\scriptsize{($\textcolor[RGB]{61,154,64} {-0.09 \%}$)} & \underline{75.12}\scriptsize{($\textcolor[RGB]{61,154,64} {-1.39 \%}$)} & \underline{21.05}\scriptsize{($\textcolor[RGB]{61,154,64} {-0.29 \%}$)}   \\

$\checkmark$   & $\checkmark$ & \textbf{39.74} & \textbf{25.82} & \textbf{18.43} & \textbf{12.72} & \textbf{15.83} & \textbf{38.24} & \textbf{76.51} & \textbf{21.34}  \\ 
\hline   
\end{tabular} 
\vspace{1mm}

\vspace{-4mm}
\label{table:ablation} 
\end{center}   
\end{table*}

\subsection{Ablation Study }
\label{sec:ablation}
\noindent\textbf{3D Compressor.}
The 3D compressor is a crucial component of 3UR-LLM, responsible for condensing 3D point cloud inputs and aligning visual and textual modalities. In previous 2D multi-modal large language models (e.g., LLaVA~\cite{liu2023improved} and BLIP-2~\cite{li2023blip-2}), the connector parameters (analogous to our 3D compressor) are typically frozen during fine-tuning to maintain strong generalization. However, as shown in Table~\ref{table:ablation}, freezing the 3D compressor's weights does not yield performance improvements and even slightly impacts the results (+1.18\% on BLEU-1, -0.49\% on BLEU-4, -0.09\% on METEOR, -0.09\% on ROUGE-L, and -1.39\% on CIDEr). These findings are consistent with those observed in 3D-LLM~\cite{hong20233d}. A possible explanation is that the pre-training data for 3D scene understanding is significantly smaller than that for 2D multi-modal tasks, necessitating the unfreezing of more parameters in 3UR-LLM to effectively learn representations and align multi-modal sequences.

\noindent\textbf{3D Query Fusion.}
The 3D query fusion module is designed to enhance the spatial perception capabilities of the 3D compressor in 3D scenes. To assess its effectiveness, we performed a component-wise analysis by retraining a model without the 3D query fusion and fine-tuning it on downstream tasks, while keeping the hyper-parameter settings consistent with our default version.
The results, presented in Table~\ref{table:ablation}, demonstrate that 3D query fusion significantly improves the model's performance, yielding gains of +1.50\% on BLEU-1, +1.17\% on BLEU-4, +0.13\% on METEOR, +0.40\% on ROUGE-L, and +2.35\% on CIDEr.

\begin{figure}[t]
\centering
\setlength{\abovecaptionskip}{0.cm}
\includegraphics[scale=0.175]{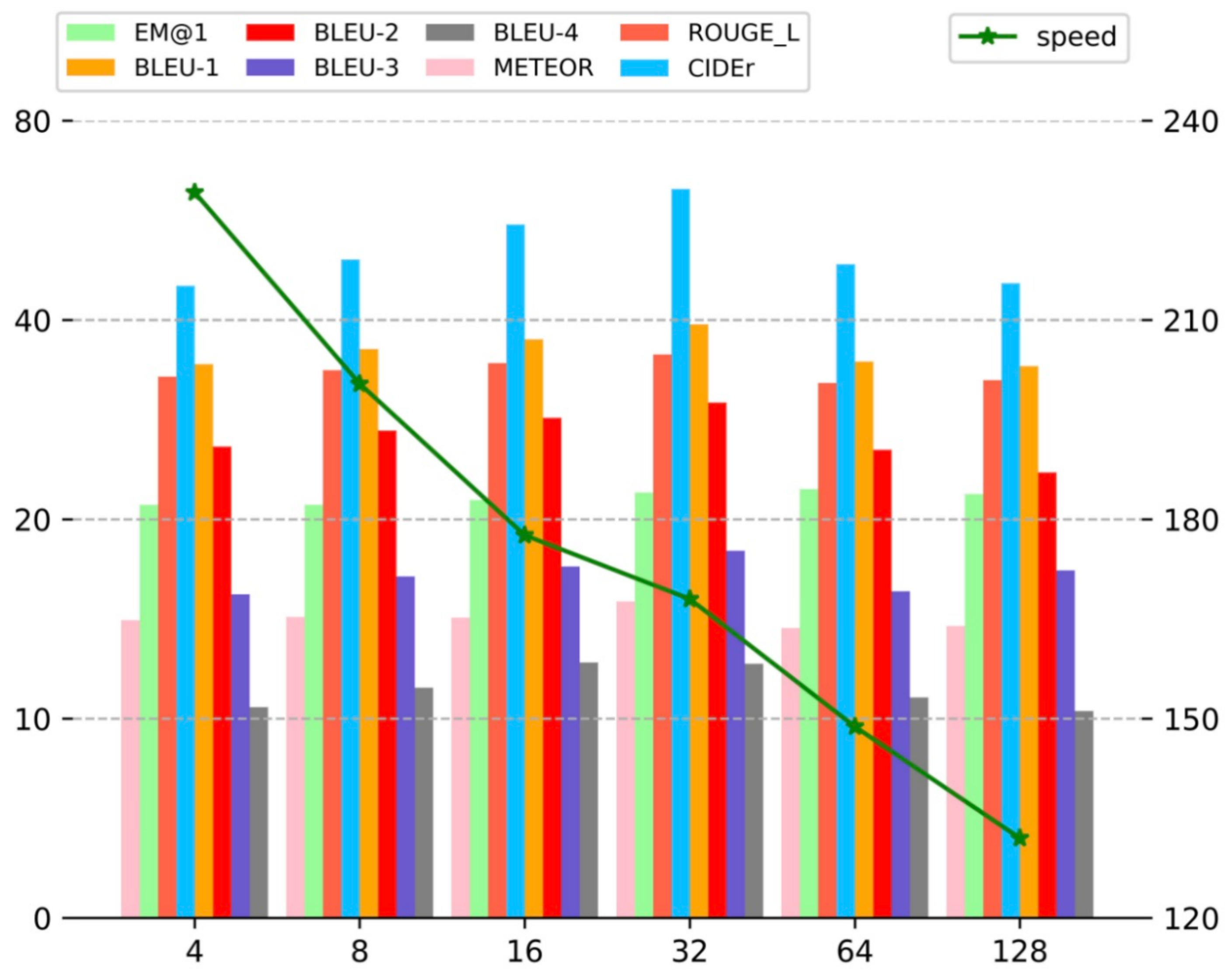}
\caption{
Ablation study on different query number which affects performance. Model inference speed measured in tokens processed per second (tokens/s).
}
\label{figure:Query}
\end{figure}

\noindent\textbf{Number of Queries.}
Optimizing the number of queries is crucial for balancing performance and efficiency. To this end, we performed ablation studies to assess the impact of varying query counts on model performance. Given the potential time overhead associated with filtering queries through 3DETR, we specifically examined the influence of query quantity on inference speed. Our findings, depicted in Fig.~\ref{figure:Query}, reveal a nuanced relationship between query count, inference speed, and performance metrics. Reducing queries to 4 enhanced the model's inference speed to 237 tokens/s but at the cost of decreased performance across several metrics: a drop of 2.01\% in BLEU-1, 2.18\% in BLEU-4, 0.95\% in METEOR, 1.12\% in ROUGE-L, and 4.86\% in CIDEr. Conversely, increasing queries to 128 diminished inference speed to 134 tokens/s and similarly led to reductions in performance metrics. A setting of 32 queries was found to deliver optimal results, achieving superior performance metrics at an acceptable speed of 186 tokens/s.


\begin{table}[h]   
\renewcommand{\arraystretch}{1.3}
\tabcolsep=0.08cm
\begin{center}   
\vspace{-2mm}
\caption{
Comparison with Q-Former in architecture on ScanQA.
Models are both pre-trained on 3DS-160K for a fair comparison.
}
\begin{tabular}{c | c c c c c c c c}   
\hline   
{Model} & {B-1} & {B-2} & {B-3} & {B-4} & {M} & {R-L} & {C} & {EM@1}\\
\hline
Ours & \textbf{36.27} & \textbf{23.47} & \textbf{16.37} & \textbf{10.97} & \textbf{14.12} & \textbf{36.90} & \textbf{70.42} & \textbf{20.02}  \\ 
cross no $\mathcal{Q}_{3D}$ & 35.04 & 23.16 & 16.19 & 10.29 & 14.07 & 36.55 & 68.18 & 19.74  \\ 
cross no $\mathcal{F}_{enc}$ & \underline{35.21} & \underline{23.29} & \underline{16.26} & \underline{10.96} & \underline{14.09} & \underline{36.87} & \underline{69.23} & \underline{19.93}  \\ 
Q-Former & {34.98} & {22.08} & {15.72} & {10.54} & {12.94} & {33.52} & {64.11} & {19.37}  \\ 
\hline   
\end{tabular} 
\label{table:same_pretrain} 
\end{center}   
\end{table}

\noindent\textbf{Efficacy of Architectural Design.}
To validate the architectural superiority of 3UR-LLM, we conducted a controlled comparison by partitioning a subset from the 3DS-160K dataset to retrain both 3UR-LLM and the previous state-of-the-art (SOTA) model, 3D-LLM~\cite{hong20233d}.
Both models employ Flan-T5-XL~\cite{chung2022scaling} as LLM backbone.
We report the experimental results in the Table~\ref{table:same_pretrain}.
According to the results, our 3UR-LLM outperforms 3D-LLM by 1.29\% on BLEU-1 (B-1), 0.43\% on BLEU-4 (B-4), 1.18\% on METEOR (M), 3.38\% on ROUGE-L (R-L), and 6.31\% on CIDEr (C). To further investigate the efficacy of our 3D compressor design, we implemented two variant models: (1) a model without 3D queries in the cross-attention mechanism, and (2) a model with the 3D feature removed from the 3D compressor, relying solely on queries from 3DETR for point cloud perception. The results indicate that removing 
$\mathcal{Q}_{3D}$ from the cross-attention significantly degrades performance, with reductions of 1.23\% in B-1 and 2.24\% in CIDEr. Similarly, eliminating  $\mathcal{F}_{enc}$ leads to performance decreases of 1.06\% in B-1 and 1.19\% in CIDEr. These findings underscore the importance of both components in our 3D compressor architecture for optimal performance.

\begin{figure*}[h]
\centering
\setlength{\abovecaptionskip}{0.cm}
\includegraphics[scale=0.2]{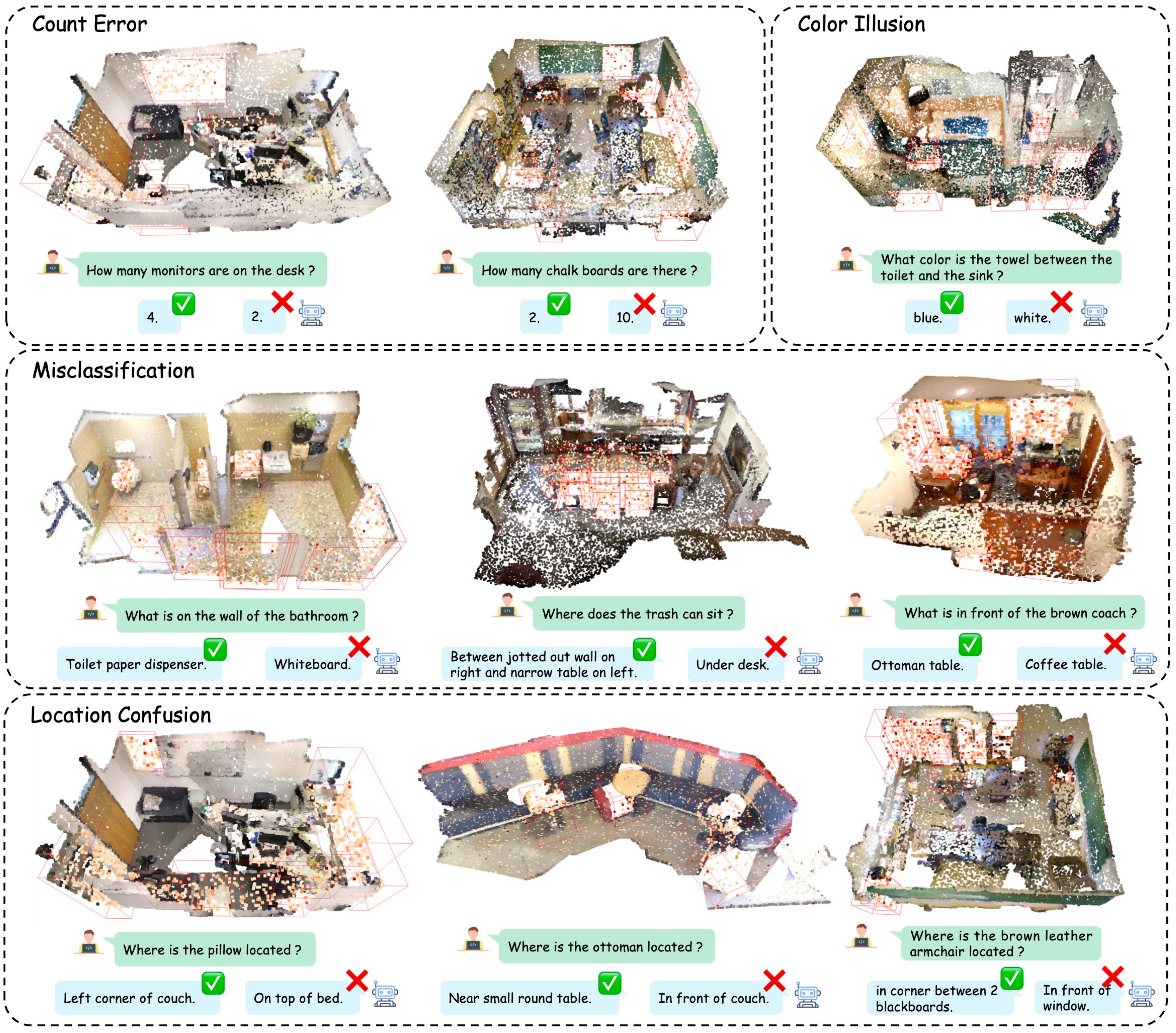}
\vspace{-0mm}
\caption{
Visualization of failure cases. 3UR-LLM demonstrates limitations in numerical assessment capabilities, recognition of color attributes, discrimination of certain categories, and the modeling of positional relationships between objects.}
\label{figure:faliure_case}
\end{figure*}

\subsection{Failure Case and Limitation} \label{failure}
Despite the commendable performance of 3UR-LLM in 3D scene comprehension, detailed visualizations of the point cloud attention process and its predictions, as shown in Fig.~\ref{figure:faliure_case}, reveal critical areas requiring enhancement. The model demonstrates a need for improved sensitivity in quantifying objects within scenes, suggesting limitations in its numerical assessment capabilities. Additionally, the accuracy in recognizing and judging color attributes within the 3D environment requires further enhancement.
Additionally, 3UR-LLM's ability to perceive the positional relationships between objects in space remains limited.
Our analysis identifies several issues: \textbf{(1) Color Illusion:} The 3D encoder's limitation is a key factor, as it primarily processes coordinates (x, y, z), restricting color recognition to positional features from $\mathcal{Q}_{3D}$ and $\mathcal{F}_{enc}$. \textbf{(2) Misclassification:} This arises from a lack of prior knowledge—information injected into the LLM lacks classification data—and redundant proposal queries that mislead the language model. \textbf{(3) Count Errors:} These stem from color illusion and misclassification issues, which impede accurate object identification and counting. \textbf{(4) Location Confusion:} This is due to insufficient pre-training data on positional relationships and inadequate position encoding, which currently relies on 2D space encoding; a new approach is needed to effectively model 3D spatial relationships.

\section{Conclusion}
In this paper, we introduce 3UR-LLM, a 3D multi-modal large language model designed to interpret complex 3D environments and generate responses to human directives. 
Distinctively, the proposed 3UR-LLM leverages point clouds as direct inputs, effectively extracting critical spatial information and mapping the condensed data into the linguistic space. 
This capability significantly enhances the model's understanding of 3D real-world scenes.  
3UR-LLM stands out through substantial enhancements in its streamlined framework, improved training efficiency, and the optimized data distribution within the newly proposed 3DS-160K dataset, all of which contribute to superior performance relative to existing methods.  Moving forward, we envision further refinements in model architecture and training protocols to accommodate even larger and more diverse 3D datasets, thereby pushing the boundaries of LLM's capability in multi-modal understanding and interaction.
\bibliographystyle{IEEEtran}
\bibliography{ref}

\vfill

\end{document}